\documentclass{article} 
\usepackage{iclr2015,times}
\usepackage{hyperref}
\usepackage{url}
\usepackage{amsmath}
\usepackage{amssymb}
\usepackage{amsthm}
\def\etal{\emph{et al }}
\iclrconference
\usepackage{graphicx}

\title{Large Scale Distributed Distance Metric Learning}

\author{
Pengtao Xie \& Eric Xing \\
School of Computer Science \\
Carnegie Mellon University \\
Pittsburgh, PA, 15213 \\
\texttt{\{pengtaox,epxing\}@cs.cmu.edu} 
}

\begin{document}

\maketitle

\begin{abstract}
In large scale machine learning and data mining problems with high feature dimensionality, the
Euclidean distance between data points can be uninformative, and Distance Metric Learning (DML)
is often desired to learn a proper similarity measure (using side information such as example data pairs being similar or dissimilar). However, high dimensionality and large volume of pairwise constraints in modern big data can lead to prohibitive computational cost for both the original DML formulation in \cite{xing2002distance} and later extensions. 
In this paper, we present a distributed algorithm for DML, and a large-scale implementation on a parameter server architecture. Our approach builds on a parallelizable reformulation of \cite{xing2002distance}, and an asynchronous stochastic gradient descent optimization procedure. To our knowledge, this is the first distributed solution to DML, and we show that, 
on a system with 256 CPU cores, our program is able to complete a DML task on a dataset with 1 million data points, 22-thousand features, and 200 million labeled data pairs, in 15 hours; and the learned metric shows great effectiveness in properly measuring distances. 
\end{abstract}

\section{Introduction}
Learning a proper distance metric \cite{xing2002distance,bilenko2004integrating, bar2005learning, globerson2005metric, weinberger2005distance, davis2007information, guillaumin2009you, kostinger2012large, mensink2012metric} is essential for many distance based data mining and machine learning algorithms, such as retrieval ~\cite{hoi2008semi}, k-means clustering \cite{xing2002distance} and k nearest neighbor (kNN) classification \cite{weinberger2005distance}. 
A commonality of these algorithms is that their accuracy depends critically on a good distance metric $M$ between points, especially when the dataset is high-dimensional. 

As first formulated in \cite{xing2002distance}, mathematically, a Distance Metric Learning (DML) problem can be formulated as semi-supervised learning problem where, given side information in the form of data pairs that are determined to be similar or dissimilar, we learn a metric $M$, which places similar data pairs close to each other, and dissimilar data pairs as far apart as possible. This leads to a quadratic program whose size grows super-linearly with the size of the data and of the side information. 
Specifically, let $M$ defines a Mahalanobis distance $(x-y)^{\mathsf{T}}M(x-y)$, where $x,y$ are $d$-dimensional feature vectors and $M\in \mathbb{R}^{d\times d}$ is a positive semidefinite matrix (to be learned). Because $M$ is a $d$-by-$d$ matrix, when the feature dimension $d$ is huge --- such as in web-scale problems where web pages are represented with bag-of-word (BOW) vectors that are millions of words long~\cite{ahmed2012scalable},
or in computer vision where hundreds of thousands of features are routinely extracted from images ~\cite{lin2011large} --- the size of $M$ quickly becomes intractable for a single machine; e.g., if $d$ contains 1 million features, then $M$ contains 1 trillion parameters. Storing this $M$ requires $\sim 4$ terabytes of memory, to say nothing of the massive computational cost of learning so many parameters. Even worse, the number of labeled data pairs can easily exceed billions or even trillions, particularly in web data. For instance, in Flickr, photos are organized into many interests groups (e.g., tiger, sushi, car) by the users --- if we simply regard photos in the same group as similar, and those in different groups as dissimilar, we can rapidly generate a huge number of similar/dissimilar pairs, as Flickr contains billions of photos and over ten million groups.

For problems involving big data volume, one would naturally speculate that modern distributed system should come for the rescue, and platforms such as Hadoop~\cite{hadoop} or Spark~\cite{zaharia2012resilient} may straightforwardly offer to scale up the originally sequential algorithms intended for a single machine. But what makes DML on large-scale data nontrivial even with popular distributed platforms is that its basic formulation requires both substantial redesign of the original algorithm, and a more optimization-friendly parallel communication strategy not supported under the currently popular bulk synchronization parallelism (BSP) adopted by Hadoop and Spark. Specifically, DML is a semi-definite programming (SDP) problem \cite{xing2002distance}, which requires eigen-decomposition of the $d$-by-$d$ Mahalanobis distance matrix $M$ after each update. Because eigendecomposition requires $O(d^3)$ computation, it is outright infeasible when the feature dimension $d$ is high, no matter how many machines are available. Furthermore, DML is a optimization problem with hard constraints (where each similar/dissimilar data pair corresponds to one constraint); this makes the distribution and parallel learning of parameters $M$ very difficult, as costly, frequent inter-machine synchronization must be employed to keep the machines' local views of $M$ consistent with each other. A BSP model would make this operation very expensive. 

Motivated by these challenges, we explore and validate new approaches to performing distributed DML on large scale problems.
On the algorithm side, inspired by ideas from \cite{weinberger2005distance}, we reformulate DML to make it tractable on high-dimensional data and amenable for distributed optimization --- specifically, we re-represent $M$ with an alternate decomposition $L^{\mathsf{T}}L$, and perform optimization directly over $L$. This not only preserves the semi-definite property of $M$, but also avoids the costly $O(d^3)$ eigen-decomposition. Moreover, to solve the inter-machine parameter synchronization challenge caused by the hard DML constraints, we use slack variables to relax those constraints, and then transform the relaxed constraints into a hinge loss function. This makes the distributed optimization much easier, yet does not hurt the effectiveness of the learned distance metric, as our validation will show.

To solve this reformulated DML problem, we build a distributed system that uses
asynchronous stochastic gradient descent to do parameter learning of $M$.
Similar and dissimilar data pairs are partitioned onto different worker machines and each worker stores a local view
of the parameter. At each iteration, each worker randomly picks up a mini-batch of data pairs,
computes a stochastic gradient and uses the gradient to update the local parameter copy.
Because the gradients computed at each worker need to be seen and utilized by other workers, we use a centralized parameter server to synchronize the parameters among workers, which has been proven to be effective both empirically and theoretically \cite{ahmed2012scalable,dean2012large, ho2013more, li2014scaling}.

The major contributions of this work are summarized as follows:
\begin{itemize}
\item We design and implement a distributed framework to support large scale distance metric learning. To our best knowledge, this is the first work targeting the parallel and distributed metric learning on large datasets.
\item We use asynchronous stochastic gradient descent to optimize DML in distributed setting and implement an efficient centralized parameter server to synchronize the parameters across machines.
\item We provide an efficient and easy-to-use implementation of the framework, that we plan to open-source.
\item We conducted distributed distance metric learning on large datasets and demonstrate the efficiency and effectiveness of our system.
\end{itemize}

In Section 2, we discuss related work, before introducing our reformulation of distance metric learning in Section 3. Section 4 presents our distributed framework for parallel distance metric learning, with experimental results in Section 5.

\section{Related Works}
Distance metric learning \cite{xing2002distance, bilenko2004integrating, bar2005learning, globerson2005metric, weinberger2005distance, davis2007information, guillaumin2009you, kostinger2012large, mensink2012metric} has been widely studied in many works. Given some data pairs labeled as similar or dissimilar, these algorithms try to learn distance metrics to make similar pairs close to each other and separate dissimilar pairs apart. The most widely used distance metric is the Mahalanobis distance. Xing \etal \cite{xing2002distance} used semidefinite programming to learn a Mahalanobis distance metric for clustering under similarity and dissimilarity constraints. They aim to minimize the distance of similar pairs while separating dissimilar pairs with a certain margin. Weinberger \etal \cite{weinberger2005distance} and Mensink \etal \cite{mensink2012metric} employed a similar semidefinite formulation for k-nearest neighbor classification. The metric is trained with the goal that the k-nearest neighbors
always belong to the same class while examples from different classes are separated by a large margin. Globerson and Roweis \cite{globerson2005metric} proposed a formulation aiming to collapse all
examples in the same class to a single point and push examples in other classes infinitely far away. Davis \etal \cite{davis2007information} proposed information theoretic metric learning which minimizes
the differential relative entropy between two multivariate Gaussians under the constraints on the distance function. 

All the above-mentioned works are designed and implemented on single machine and cannot handle large scale data efficiently. Kostinger \etal \cite{kostinger2012large} proposed a fast DML method based on likelihood-ratio test, which does not require iterative optimization procedures and hence is very efficient and scalable. However, distance metrics learned by this method yield poor performance empirically and this method is less flexible (can only deal with equivalence constraints). Our method learns effective distance metrics on multiple datasets and can flexibly deal with not only pair-wise constraints but also triple-wise constraints. In terms of the size of the problem, Mensink \etal \cite{mensink2012metric} achieved similar scale on ImageNet ~\cite{deng2009imagenet} dataset as we do. However, in their work, DML was performed on a single machine, which takes days to finish. In our work, with the distributed framework, we can finish learning within 15 hours.

Designing and implementing parameter servers (PS) for distributed machine learning have been explored in several works \cite{ahmed2012scalable, dean2012large, ho2013more, li2014scaling} . Most of these designs contain a centralized server and a collection of workers. The central server maintains the global parameter and each worker has a local copy of the parameter. Workers compute local updates of the parameter and push the updates to the central server. The central server aggregates updates from workers, apply the update to the global parameter and push the fresh global parameter back to workers. The existing parameter servers differ in consistency control. In Bulk Synchronous Parallel (BSP) systems like Hadoop \cite{dean2008mapreduce}, Spark \cite{zaharia2012resilient}, workers must wait for each other at the end of every iteration. In asynchronous parameter server \cite{ahmed2012scalable, dean2008mapreduce}, all workers work on their own pace and never waits. Staleness Synchronous Parallel \cite{ho2013more} seeks a balance between BSP and ASP. Workers are allowed to see different versions of the parameter, but the difference is bounded. None of the above works has studied distance metric learning (DML) before and it is unclear whether the parameter server based framework can be effective for distributed DML.


\section{Distance Metric Learning}
In the original formulations \cite{xing2002distance, globerson2005metric, weinberger2005distance} of distance metric learning, a semi-definite programming problem (SDP) needs to be solved. SDP requires eigen-decomposition of the Mahalanobis distance matrix, which is very hard to do on high dimensional data, if not impossible. 
To make DML scalable on large scale problems, similar to \cite{weinberger2005distance}, we do the reformulations in two ways. First, we factorize the Mahalanobis matrix $M$ into $M=L^{\mathbf{T}}L$ and do the optimization over $L$ instead of $M$. The factorization ensures $M$ still to be positive semi-definite, but avoids the eigen-decomposition of $M$. Second, we use slack variable to relax the constraints over dissimilar pairs and use hinge loss to replace the constraints. This can turn the original constrained problem into an unconstrained problem which makes its distributed optimization much easier without sacrificing the quality of learned distance metric. 

\subsection{Reformulation of Distance Metric Learning}
Distance metric learning is a family of algorithms and has various formulations regarding the metric to learn, the form of distance supervision and how the objective function is defined. Among them, the most popular setting is: 1, metric: Mahalanobis distance $(x-y)^{\mathsf{T}}M(x-y)$, where $x$ and $y$ are data samples and $M$ is a symmetric and positive semidefinite matrix to be learned; 2, the form of distance supervision: pairs of data samples either labeled as similar or dissimilar; 3, learning objective: learn a distance metric to place similar points as close as possible and dissimilar points as far as possible. Given a set of pairs labeled as similar $\mathcal{S}=\{(x_{i},y_{i})\}_{i=1}^{|\mathcal{S}|}$ and a set of pairs labeled dissimilar $\mathcal{D}=\{(x_{i},y_{i})\}_{i=1}^{|\mathcal{D}|}$, DML learns the Mahalanobis distance by optimizing the following problem
\begin{equation}
\label{eq:dml_opt_1}
\begin{array}{ll}
\textrm{min}_{M}&\sum\limits_{(x,y)\in \mathcal{S}}(x-y)^{\mathsf{T}}M(x-y)\\
s.t.&(x-y)^{\mathsf{T}}M(x-y)\geq1, \forall (x,y)\in\mathcal{D}\\
&M\succeq 0
\end{array}
\end{equation}
where $M\succeq 0$ denotes that $M$ is required to be positive semidefinite. This optimization problems tries to minimize the Mahalanobis distances between all pairs labeled as similar while separating dissimilar pairs with a margin 1. $M$ is required to be positive semidefinite to ensure that Mahalanobis distance is a valid metric. 

While this problem can be solved using projected gradient descent (PGD), it turns out to be infeasible for high dimensional problems. In each iteration of PGD, one first computes the gradient of $M$ w.r.t the objective function $\sum_{(x,y)\in \mathcal{S}}(x-y)^{\mathsf{T}}M(x-y)$, then updates $M$ using gradient descent and finally project $M$ onto the convex set specified by the two constraints. Doing projection requires the eigen-decomposition of $M$, whose complexity is $O(d^{3})$, where $d$ is the feature dimension of the data points. For high dimension problems where $d$ can be hundreds of thousands or even millions, eigen-decomposition is extremely hard and inefficient, if not impossible. Note that a symmetric and positive semidefinite matrix $M$ can always be factorized into $M=L^{\mathsf{T}}L$ \cite{weinberger2005distance}, where $L$ is a matrix of size $k\times d$ and $k\leq d$. Replacing $M$ with $L^{\mathsf{T}}L$, the problem defined in Eq.(\ref{eq:dml_opt_1}) can be written as
\begin{equation}
\label{eq:dml_opt_2}
\begin{array}{ll}
\textrm{min}_{L}&\sum\limits_{(x,y)\in \mathcal{S}}\|L(x-y)\|^{2}\\
s.t.&\|L(x-y)\|^{2}\geq1, \forall (x,y)\in\mathcal{D}\\
\end{array}
\end{equation}
The constraint requires the distance between dissimilar pairs to be greater than a margin of 1, which is hard to enforce in the distributed setting where communication between machines is limited. We adopt a strategy similar to \cite{weinberger2005distance}, which introduces slack variables $\xi$ to relax the hard constraint in Eq.(\ref{eq:dml_opt_2}) and get 
\begin{equation}
\label{eq:dml_opt_3}
\begin{array}{ll}
\textrm{min}_{L}&\sum\limits_{(x,y)\in \mathcal{S}}\|L(x-y)\|^{2}+\lambda \sum\limits_{(x,y)\in \mathcal{D}}\xi_{x,y}\\
s.t.&\|L(x-y)\|^{2}\geq1-\xi_{x,y}, \xi_{x,y}\geq 0, \forall (x,y)\in\mathcal{D}\\
\end{array}
\end{equation}
Using hinge loss, the constraint in Eq.(\ref{eq:dml_opt_3}) can be further eliminated
\begin{equation}
\label{eq:dml_opt_4}
\begin{array}{ll}
\textrm{min}_{L}\sum\limits_{(x,y)\in \mathcal{S}}\|L(x-y)\|^{2}
+\lambda \sum\limits_{(x,y)\in \mathcal{D}}\textrm{max}(0, 1-\|L(x-y)\|^{2})\\
\end{array}
\end{equation}
To this end, we turn the original constrained problem into an unconstrained one and the optimization becomes much easier. We use stochastic gradient method to do optimization. 

\section{Distributed Parallel Distance Metric Learning}
In large scale distance metric learning problems, the number of similar and dissimilar pairs can be tens of millions. The dimensionality of features can be hundreds of thousands. Learning distance metrics on such large datasets on a single machine is infeasible. To solve this problem, we design a distributed framework to do parallel DML. 
We use asynchronous stochastic gradient descent to do the optimization. We partition the similar pairs and dissimilar pairs onto different machines. Each machine uses the data pairs it holds to update the model parameters in an asynchronous manner. Parameters on different machines are synchronized through a centralized parameter server. In terms of the form of distance supervision, we focus on pair-wise constraints (e.g., $i$ is similar to $j$; $j$ is dissimilar to $k$) in this work. Our framework can be easily extended to support triple-wise constraints \cite{weinberger2005distance} (e.g., $i$ is more similar to $j$ than to $k$).

\subsection{PARAMETER SYNCHRONIZATION}
\begin{figure}
\begin{center}
\includegraphics[width=0.3\columnwidth]{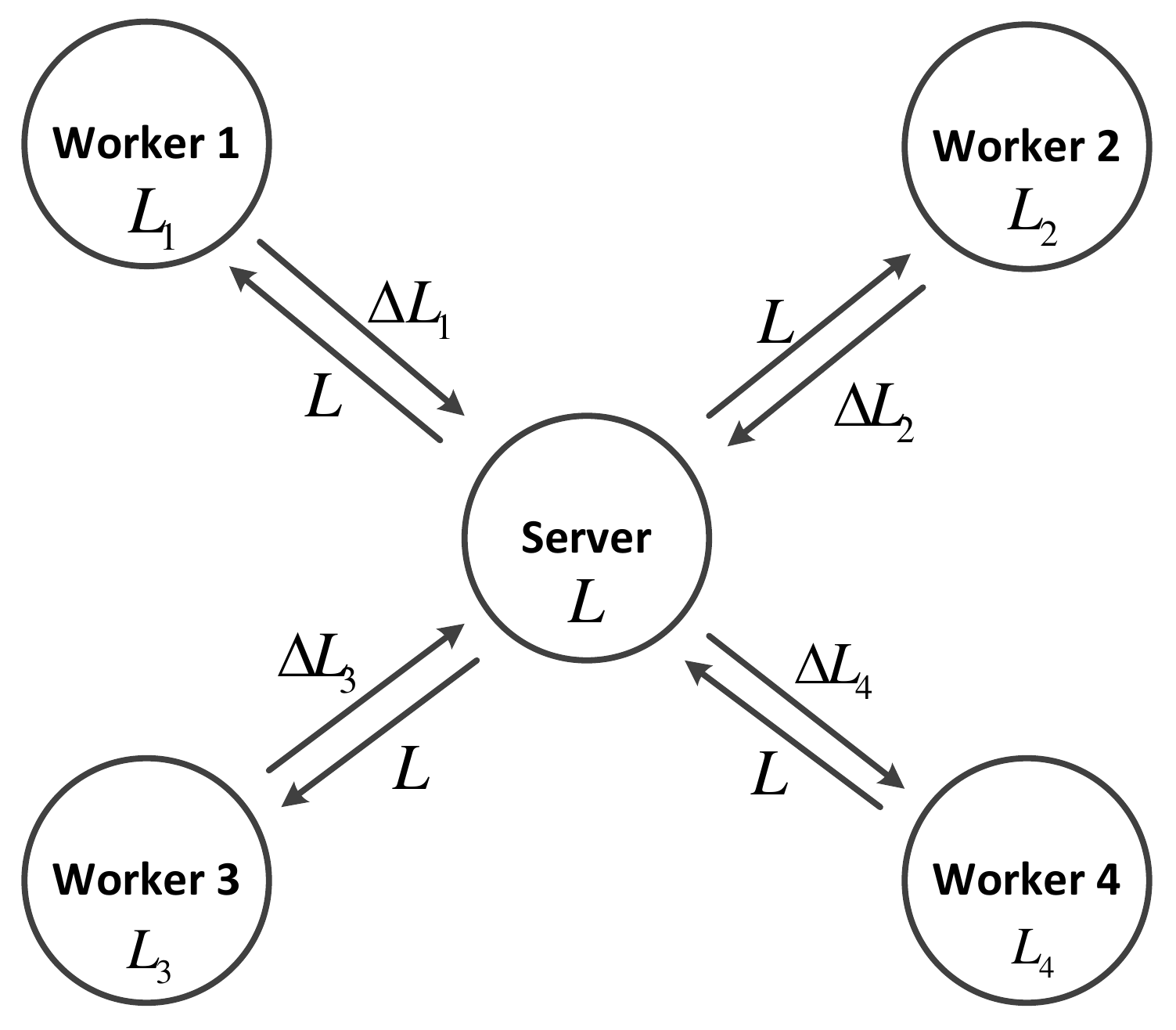}
\includegraphics[width=0.6\columnwidth]{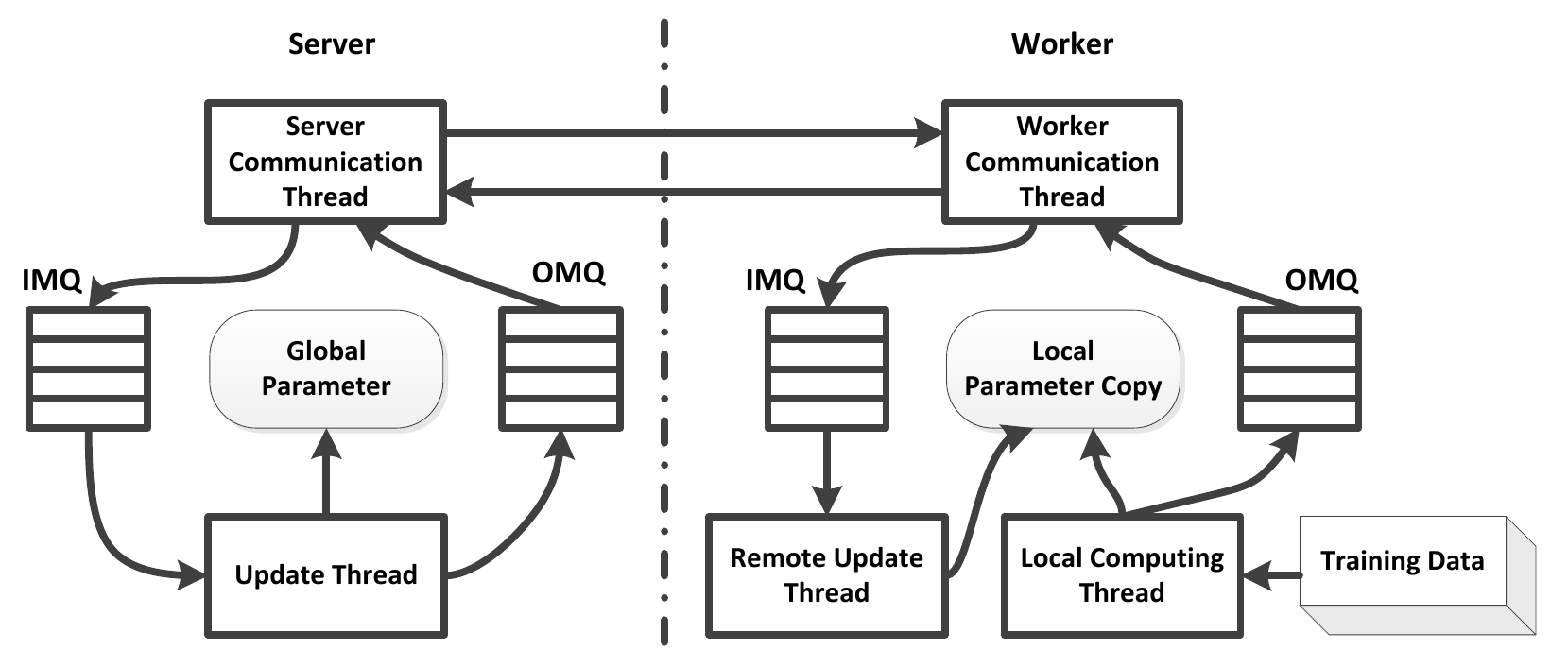}
\caption{Logical View of Parameter Server. 
A centralized server maintains the global parameter $L$. Each work $p$ has a local copy $L_{p}$ of the global parameter. In each iteration, each work takes a data pair and computes a gradient update $\bigtriangleup L_{p}$ and sends $\bigtriangleup L_{p}$ to the central server. The server aggregates the gradients received from workers and uses them to update the global parameter $L$. $L$ is then pushed back to each worker and workers replace their local copy with $L$.
}
\end{center}
\end{figure}

To learn $L$ in a distributed environment with $P$ worker machines, we partition the similarity pair $\mathcal{S}$ and dissimilar pair $\mathcal{D}$ into $P$ pieces $\mathcal{S}_{1}$,$\mathcal{S}_{2}$,$\cdots$,$\mathcal{S}_{P}$ and $\mathcal{D}_{1}$,$\mathcal{D}_{2}$,$\cdots$,$\mathcal{D}_{P}$ and each machine holds one piece. Each machine $p$ has a local copy $L_{p}$ of the global parameter $L$. And different parameter copies are synchronized across machines to ensure they are as much the same as possible. Similar to \cite{ahmed2012scalable, dean2008mapreduce, ho2013more, lee2014primitives}, we use a centralized parameter servers to synchronize these parameter copies. The central server holds the global parameter $L$. 
Each worker communicates with the central server and workers do not communicate with each other. At each iteration of the stochastic gradient descent algorithm, each worker $p$ randomly samples a minibatch of data pairs from both the similar pair set $\mathcal{S}_{p}$ and the dissimilar pair set $\mathcal{D}_{p}$ it holds and computes a gradient update $\bigtriangleup L_{p}$ using the local parameter copy $L_{p}$ and the minibatch. Worker $p$ sends $\bigtriangleup L$ to the central server. The central server aggregates updates received from workers and use them to update the central parameter $L$. The central server sends the updated $L$ to each worker and the worker replaces its local parameter copy $L_{p}$ with $L$.

\subsection{Implementation Details}


The centralized parameter server contains one central server and $P$ workers. Workers communicate with the central server and they do not communicate with each other. The server maintains the global parameter and each worker has a local copy of the parameter. 
On the server side, there are two threads: 1) update thread; 2) communication thread. The server maintains two message queues to exchange messages with workers: 1) inbound message queue; 2) outbound message queue.
The communication thread receives gradient updates from workers and puts them into the inbound message queue. The update thread takes a batch of gradient updates from the inbound message queue and uses them to update the central parameter. Then the update thread puts the updated central parameter to the outbound message queue. The communication thread takes updated parameter out of the outbound message queue and sends it to all workers.

On the worker side, there are three threads: 1) local computing thread; 2) remote update thread; 3) communication thread. The worker also maintains two message queues: 1) inbound message queue; 2) outbound message queue. At each iteration, the local computing thread takes a minibatch of data pairs, computes the gradient, uses the gradient to update the local parameter copy and puts the gradient into the outbound message queue. The communication thread sends the gradients in the outbound message queue to the server. On the other hand, it receives fresh parameters from the server and puts them into the inbound message queue. The remote update thread takes parameters out of the inbound message queue and uses them to replace the local parameter copy on this worker. 

The server threads and worker threads execute in a best-effort manner, which means whenever some work is available to do, the threads do it without caring about other threads. For example, whenever there is a message in the outbound message queue, the communication thread sends them out. Whenever there is a message in the inbound message queue, the update thread uses it to update the parameter. Each thread behaves as if no other threads exist. The threads are coordinated indirectly by the message queues.

\section{Experiments}
\begin{table*}[t]
\caption{Statistics of Datasets}
\label{table:dataset}
\begin{center}
\begin{tabular}{c|c|c|c|c|c|c}
\hline
Dataset& feat. dim & $k$& \# parameters &\#samples & \#similar pairs &\#dissimilar pairs\\
\hline
MNIST& 780 &600 & 0.47M & 60K & 100K &100K\\
ImNet-60K& 21504 &10000 & 220M & 63K & 100K &100K\\
ImNet-1M& 21504 &1000 & 21.5M & 1M & 100M &100M\\
\hline
\end{tabular}
\end{center}
\vskip -0.2in
\end{table*} 
\subsection{Datasets}
We use three datasets in our experiments. Table \ref{table:dataset} summarizes the statistics of these datasets. The first one is the MNIST hidden written digits ~\cite{lecun1998gradient}. It has 60K training images and 10K testing images. Each image is of size $32\times 32$ and has a digit label. Images are represented with raw pixels, with a dimensionality of 780.
We randomly sample 100K ``similar'' pairs and 100K ``dissimilar'' pairs from the training images. If two images are from the same digit, we label them as ``similar''. If two images are from different digits, we label them as ``dissimilar''. 

The second dataset we use is ImageNet-63K, which has 63K training images randomly selected from the ImageNet ~\cite{deng2009imagenet} dataset. Each image is associated with a class label and the total number of classes is 1000. Images are represented with Locality-constrained Linear Coding (LLC) \cite{wang2010locality}, with a dimensionality of 21504.
We randomly sample 100K ``similar'' pairs and 100K ``dissimilar'' pairs from the training images. If two images are from the same class, we label them as ``similar''. If two images are from different classes, we label them as ``dissimilar''. 

The third dataset is ImageNet-1M. It has 1 million training images and 63K testing images randomly selected from ImageNet~\cite{deng2009imagenet}. The total number of classes is 1000. Images are represented with LLC features. We randomly sample 100M ``similar'' pairs and 100M ``dissimilar'' pairs to learn the distance metric.

\subsection{Experimental Setup}
In all the experiments, the tradeoff parameter $\lambda$ is set to 1 and the threshold $c$ is set to 1.
For MNIST, we set $k$ (the number of rows of $L$) to 600.  At each iteration, we use a minibatch of 1000 pairs (500 similar pairs and 500 dissimilar pairs) to do the update.
For ImageNet-63K, we set $k$ to 10000. The number of model parameters is about 220 million. At each iteration, we use a mini-batch of 100 pairs (50 similar pairs and 50 dissimilar pairs) to do the update. 
For ImageNet-1M, we set $k$ to 1000 to make the computation tractable. The mini-batch size is 1000 (500 similar pairs and 500 dissimilar pairs). Experiments on MNIST are done on machines each of which has 16 CPUs and 64G main memory. Experiments on ImageNet-63K and ImageNet-1M are performed on machines each of which has 64 CPUs and 128G main memory.

\subsection{Scalability of Our Framework}





\begin{figure}[t]
\begin{center}
\includegraphics[width=0.3\columnwidth]{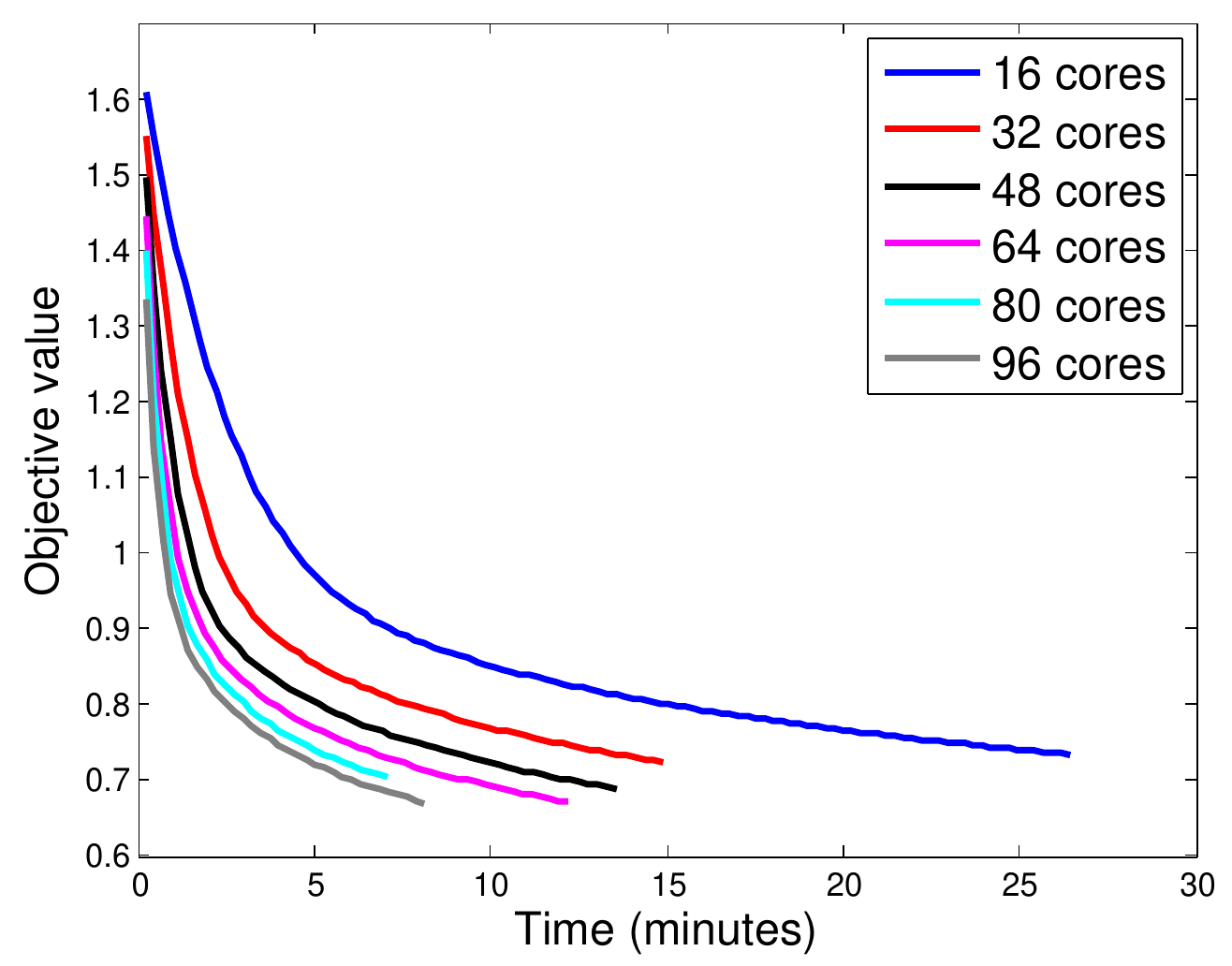}
\includegraphics[width=0.3\columnwidth]{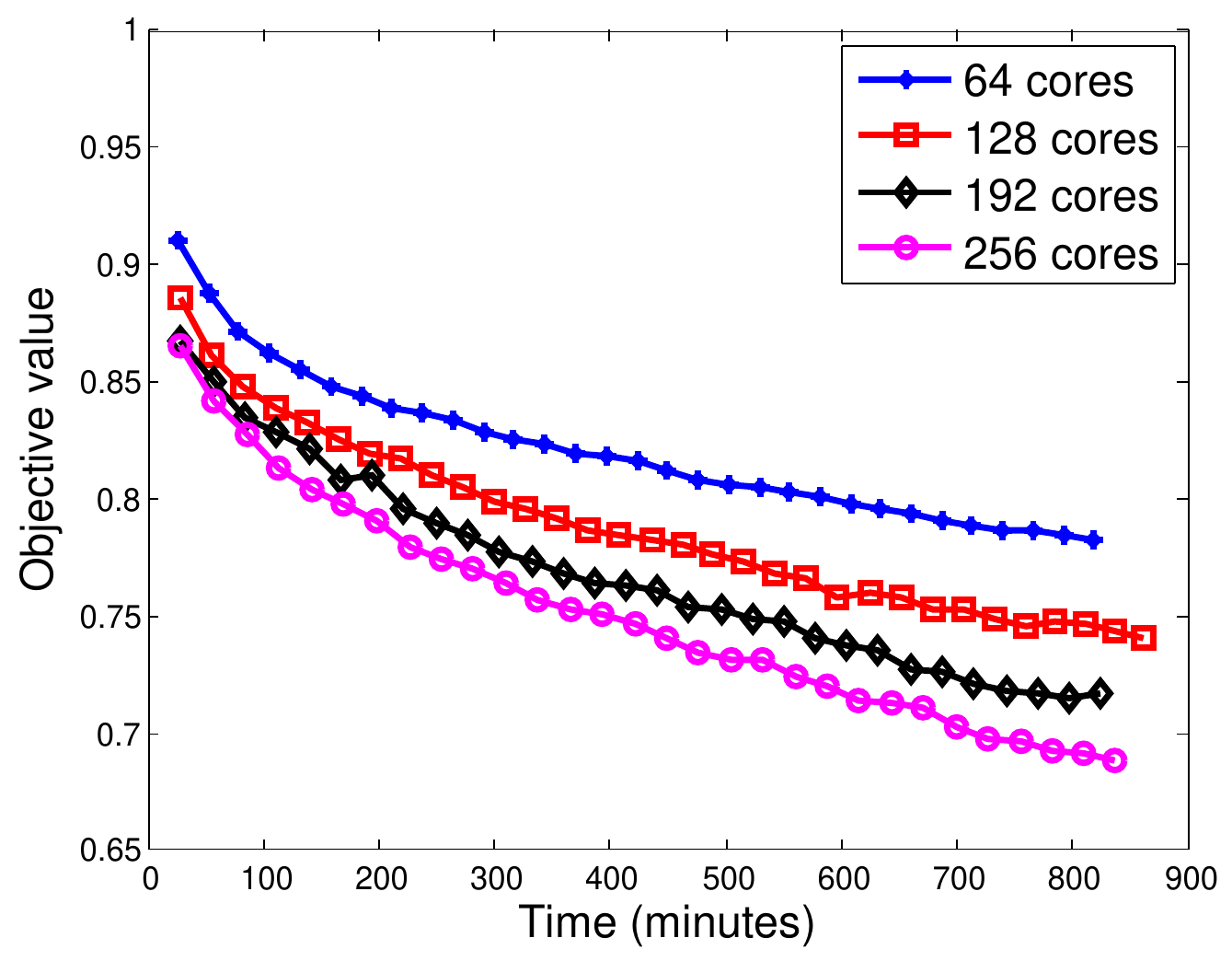}
\includegraphics[width=0.3\columnwidth]{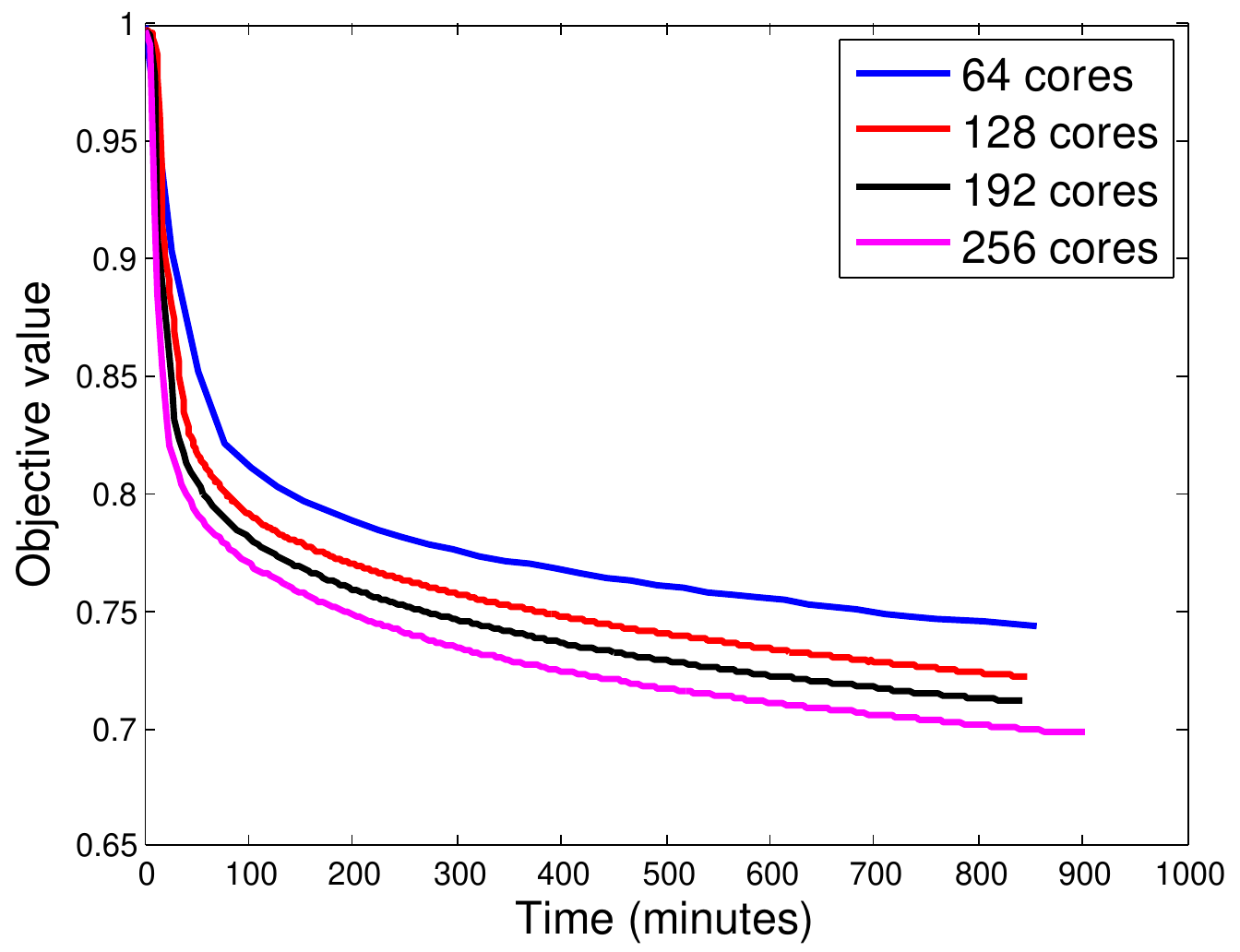}
\caption{(a) Convergence curves on MNIST dataset. (b) Convergence curves on ImageNet-63K dataset. (c) Convergence curves on ImageNet-1M dataset.}
\label{fig:cvg}
\end{center}
\vspace{-2 mm}
\end{figure}

We evaluated the scalability of our framework w.r.t the number of CPU cores on three datasets. 
Figure \ref{fig:cvg}(a) shows the convergence curves on MNIST dataset under different number of CPU cores. The horizontal axis corresponds to running time measured in minutes. The vertical axis corresponds to objective values. Figure \ref{fig:cvg}(b) and \ref{fig:cvg}(c) show the convergence curves on ImageNet-63K and ImageNet-1M respectively.
As can be seen from the three figures, increasing the number of machines consistently increases the convergence speed. 

\begin{figure}[t]
\begin{center}
\includegraphics[width=0.3\columnwidth]{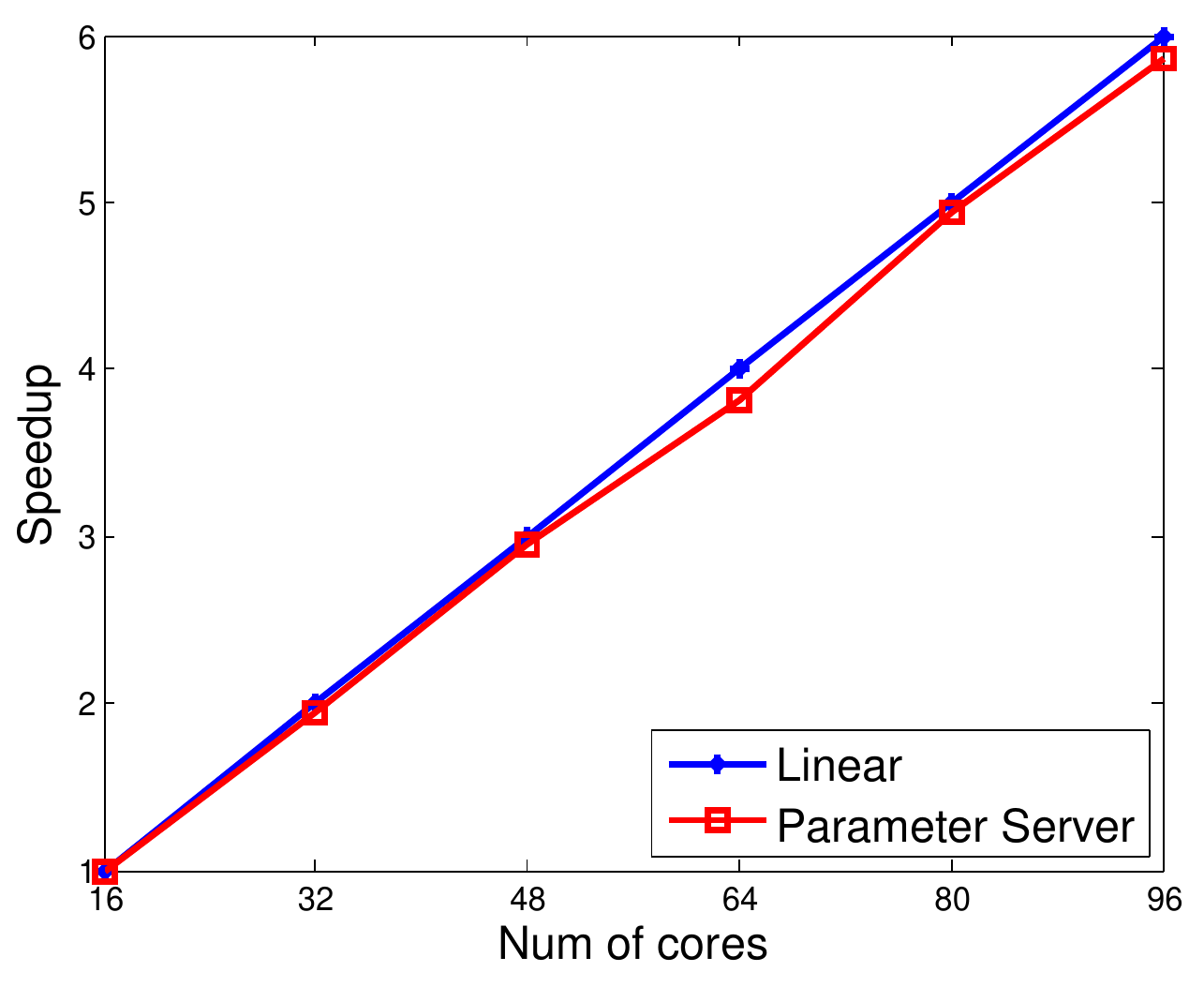}
\includegraphics[width=0.3\columnwidth]{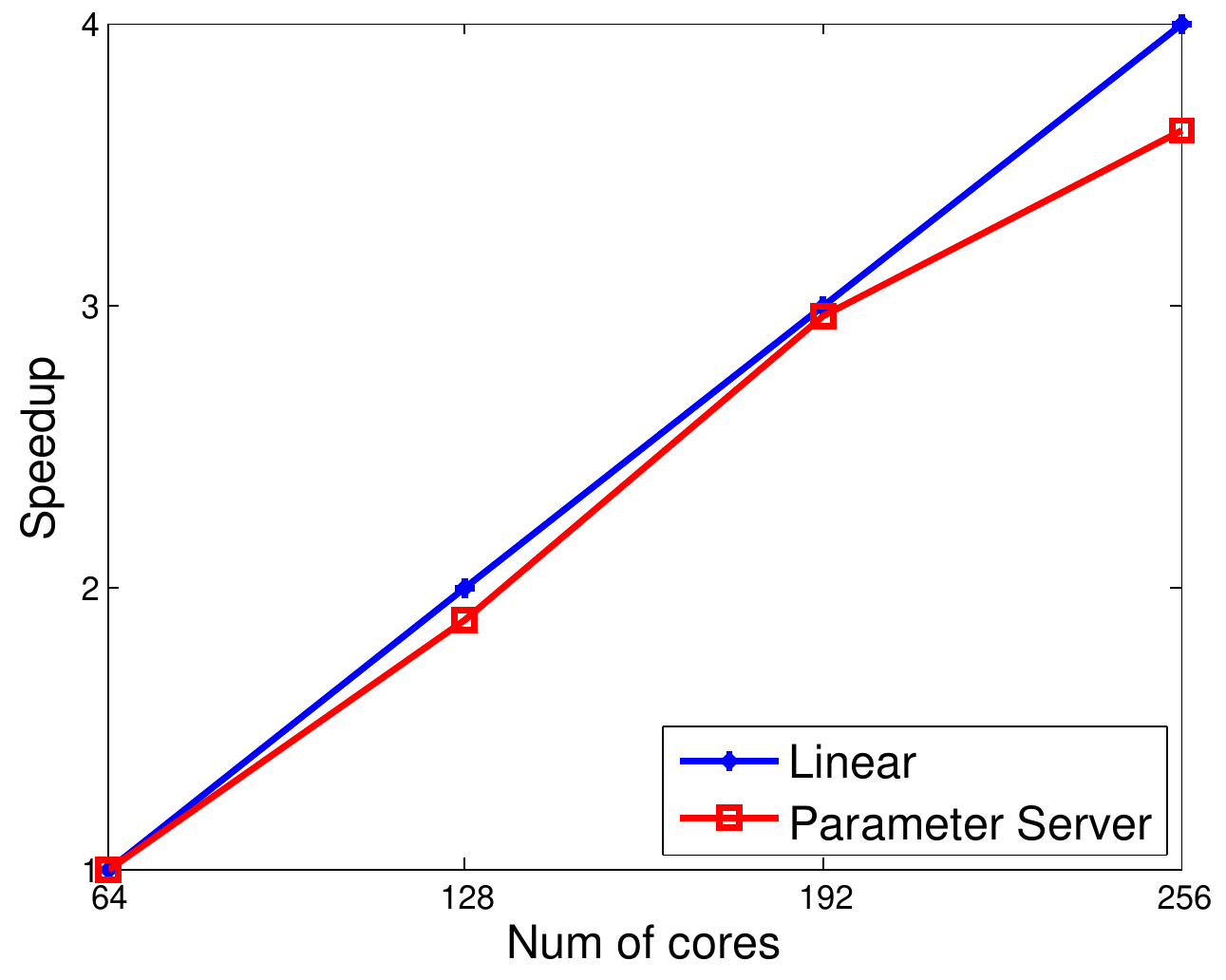}
\includegraphics[width=0.3\columnwidth]{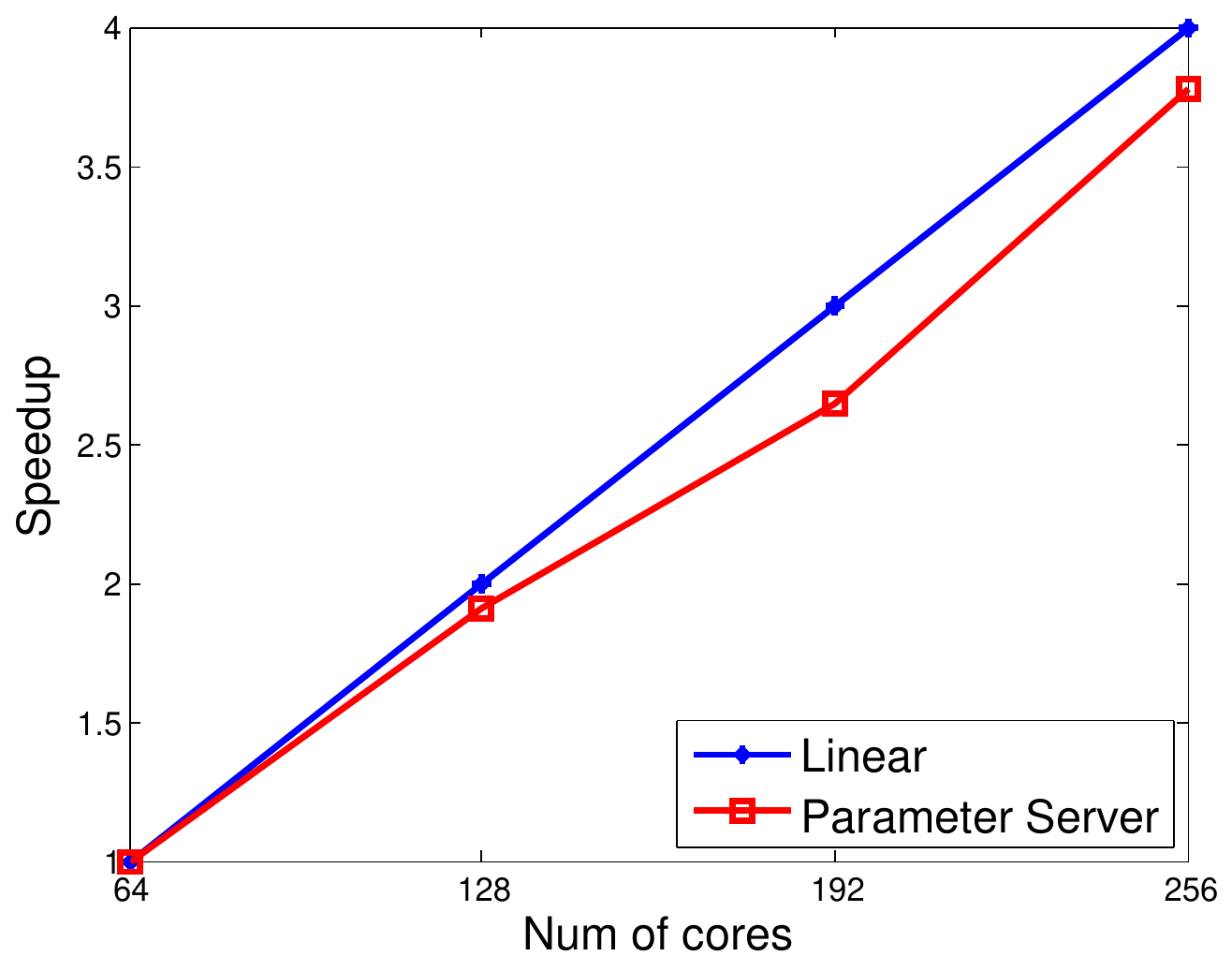}
\caption{(a) Speedup on MNIST dataset. (b) Speedup on ImageNet-63K dataset. (c) Speedup on ImageNet-1M dataset.}
\label{fig:spdup}
\end{center}
\vspace{-5 mm}
\end{figure}

To measure how our framework can speedup the training as we increase the number of machines (cores), for each machine setting we record the running time that the objective value is decreased to $p$, where $p$ is the objective value achieved by one single machine at the end of training. The speedup factor of $n$ machines is calculated as $t_{n}/t_{1}$, where $t_{n}$ is the running time of $n$ machines and $t_{1}$ is the runtime of 1 machine.
Figure \ref{fig:spdup}(a) shows the speedup factors on MNIST dataset. The horizontal axis corresponds to the number of cores. The vertical axis shows speed up factors. The blue curve is linear speedup (optimal speedup). The red curve shows the speed up factors achieved by our framework. As can be seen from this figure, our framework achieves near optimal speedup under all machine settings with different CPU cores.
Figure \ref{fig:spdup}(b) and \ref{fig:spdup}(c) shows the speedup on ImageNet-63K and ImageNet-1M dataset respectively. Our framework achieves speedups which are very close to linear under different number of cores. With 4 machines (256 cores in total), our framework achieves 3.6 times speedup on ImageNet-63K dataset and achieves 3.8 times speedup on ImageNet-1M dataset compared with those on 1 machine (16 cores in total). This demonstrates that our framework scales very well with the number of CPU cores (machines). 

The scalability of our framework attributes to the sufficient use of CPU cores to do computation and efficient parameter synchronization among workers. We distribute the computation over all available machines and allocate a copy of the model parameters to each machine. Computing threads on each machine always have access to the parameters through the parameter copy, thereby, computation can be ceaselessly performed without blocking caused by the coordination and communication with other machines. Each machine runs as if no other machines exist. This ensures that all CPU cores can be sufficiently used. On the other hand, we synchronize the parameter copies of different workers using a centralized server, to ensure that each work's update can be timely contributed to the global parameters and the computing threads on each worker can use the most-up-to-date parameters to do computation.
The synchronization happens in background and does not require blocking computation.

%

\subsection{Quality of the Learned Distance Metric}




\begin{figure}[t]
\begin{center}
\includegraphics[width=0.3\columnwidth]{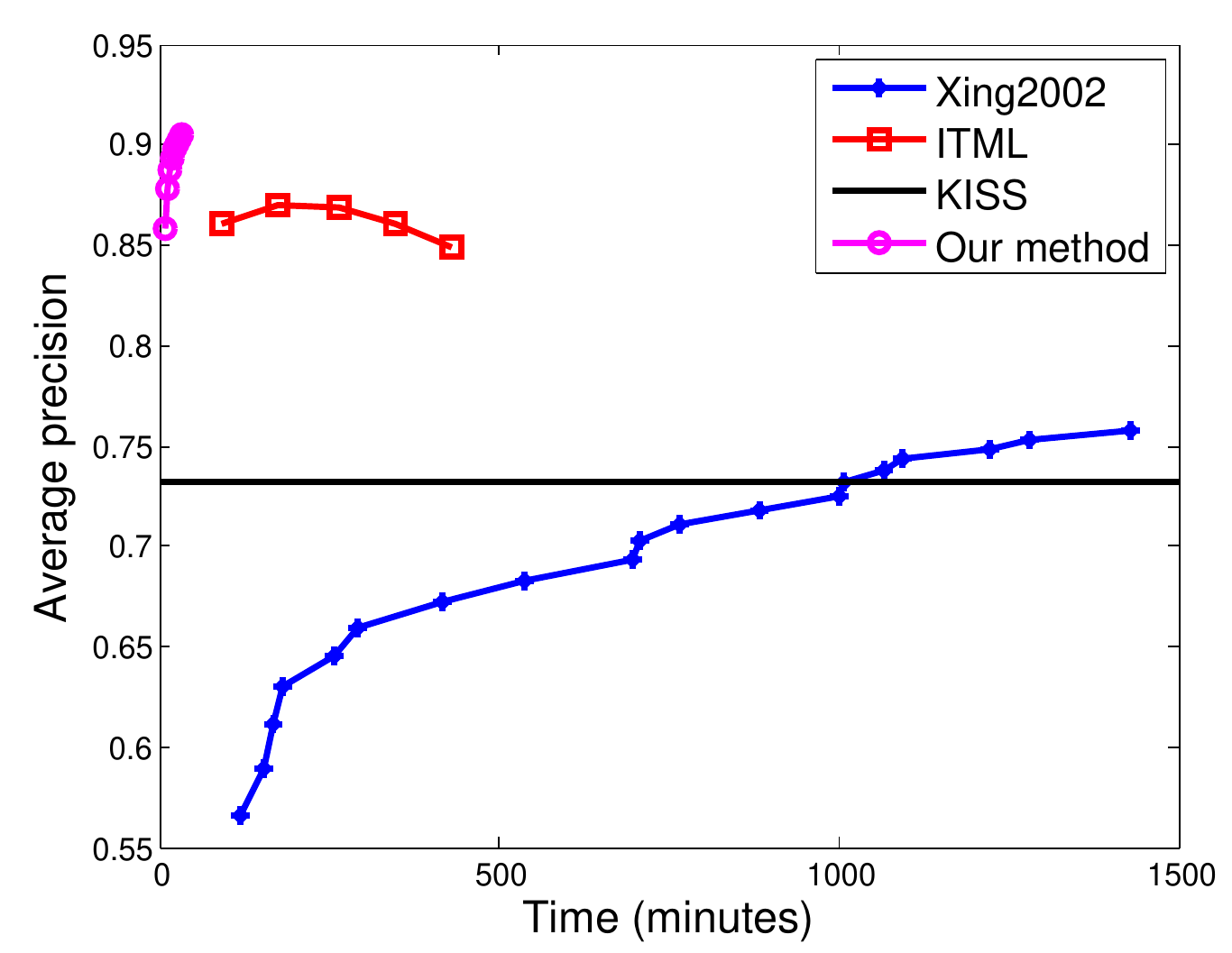}
\includegraphics[width=0.3\columnwidth]{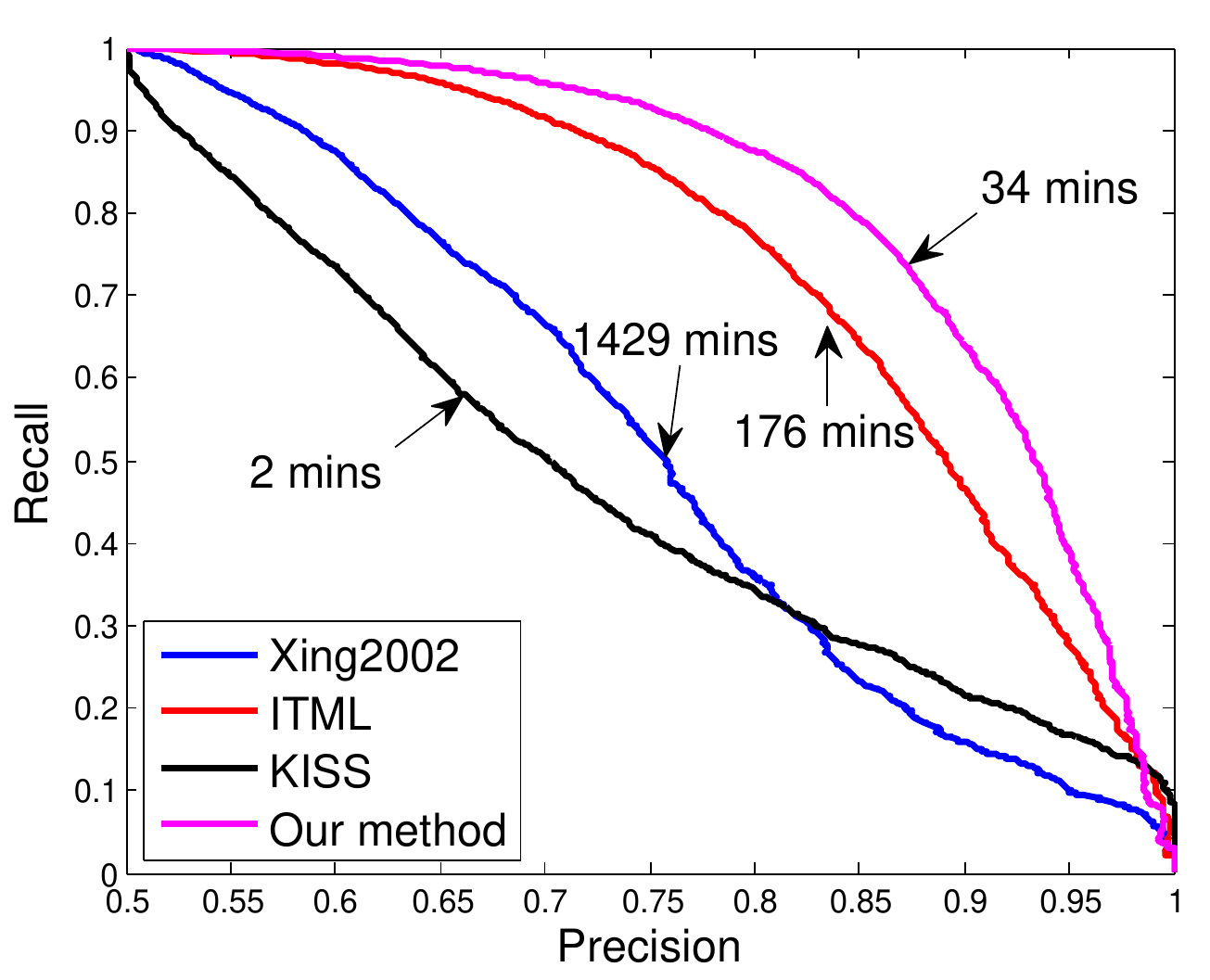}
\includegraphics[width=0.3\columnwidth]{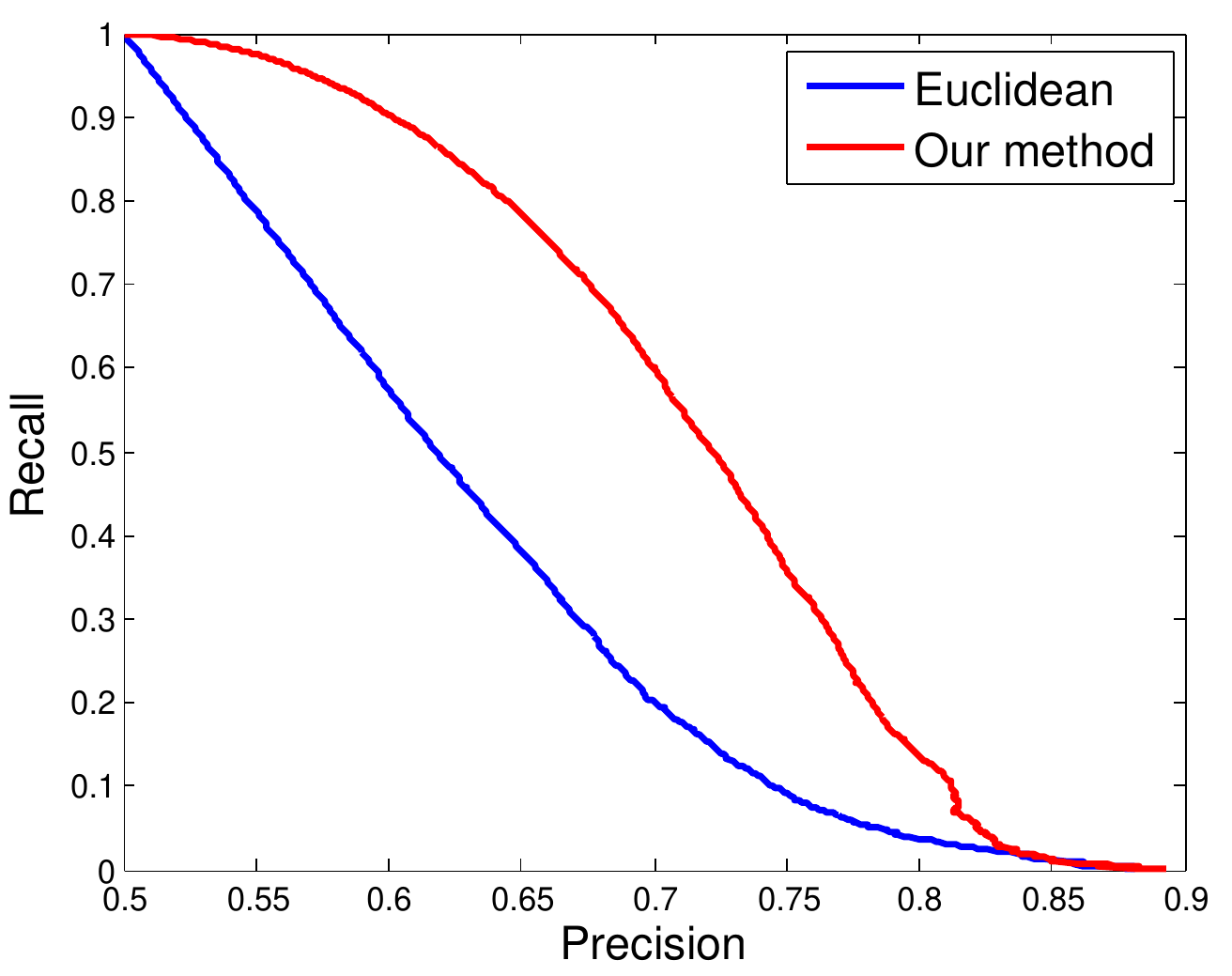}
\caption{(a) Average precision versus running time on MNIST dataset. (b) Precision-recall curves on MNIST dataset. (c) Precision-recall curves on ImageNet-1M dataset.}
\label{fig:pfm}
\end{center}
\vspace{-6.5 mm}
\end{figure}

We also measure the quality of the learned distance metric.
 We compare our reformulation in Eq.(\ref{eq:dml_opt_4}) with the original formulation \cite{xing2002distance} in Eq.(\ref{eq:dml_opt_1}) (denoted by Xing2012) and with Information Theoretical Metric Learning (ITML) \cite{davis2007information} and the likelihood test based method (KISS) \cite{kostinger2012large}.  All methods (including our method) are implemented with MATLAB and runs on a single thread. For each method, we learn a distance metric on the MNIST training set. In ITML, we set the tradeoff parameter $\gamma$ is set to 0.001. In KISS, the feature dimension is reduced to 600 using Principal Component Analysis (PCA) to ensure the covariance matrices are invertible. 

 To evaluate the effectiveness of the learned metric, we randomly sample 10K similar pairs and 10K dissimilar pairs from 100K held-out testing images and use the metric to judge whether these pairs are similar or dissimilar. If the distance is great that some threshold $t$, the pair is regarded as similar. Otherwise, the pair is regarded as dissimilar. We use two evaluation metrics: average precision and precision-recall curves. 
 
 Figure \ref{fig:pfm}(a) shows the average precision versus running time. Figure \ref{fig:pfm}(b) shows the best precision-recall curve achieved by each method.  As can be seen from the two figures, our method is not only much faster, but also achieves better performance. 
Optimizing the original formulation in Eq.(\ref{eq:dml_opt_1}) (Xing2002) requires solving a linear constrained least square (LCLS) problem at each iteration, which is very costly. On the MNIST dataset, the number of variables to be optimized in LCLS is 680K and the number of constraints is 100K. Solving such a LCLS problem in each iteration is highly demanding.  
Besides, the original formulation requires eigen-decomposition of the Mahalanobis matrix $M$, which is very costly when the feature dimension is high. In ITML, the computational complexity on each data pairs is $O(d^{2})$, where $d$ is the feature dimension. In our method, the complexity is $O(dk)$, where $k$ is usually smaller than $d$. 
From Figure \ref{fig:pfm}(a), we observe that in ITML, the precision is not consistently increasing as running time increases. This is because in each iteration of ITML, a single data pair is used to update the parameter, which may incur high variance. It is unclear how to extent ITML to mini-batch update to make the algorithm stable. KISS computes a distance metric in one shot and does not require iterative optimization, thereby, it is very fast. It learned a metric in 2 minutes. However, the metric yields very pool performance. The average precision is only 0.73 (Figure \ref{fig:pfm}(a)) and the precision-recall curve (Figure \ref{fig:pfm}(b)) is much worse than other methods. Our method is very efficient. The training is finished in about half a hour on a single thread while Xing2002 takes about 24 hours and ITML takes about 3 hours. With our distributed framework, the speed can be further improved significantly as shown in Figure \ref{fig:spdup}(a).
In addition, our method is very effective. It achieves an average precision of 0.90, which cannot be achieved by the other three methods.

We also evaluate the effectiveness of the distance metric learned on ImageNet-1M. From 63K held-out testing images, we randomly sample 100K similar pairs and 100K dissimilar pairs and compute precision-recall curves, which are shown in Figure \ref{fig:pfm}(c). The blue curve is obtained from Euclidean distance computed on original feature vectors. The red curve corresponds to Mahalanobis distance computed under the learned distance metric. As can be seen from the figure, with distance metric learning, the performance is greatly improved.

\section{Conclusions}
\vspace{-2 mm}
In this paper, we reformulate the original DML problem into an unconstrained optimization problem that is amenable for parallelization, and use it as the basis for a distributed framework to support large scale distance metric learning.
Our framework makes use of asynchronous stochastic gradient descent for distributed optimization, where the computation is distributed across worker machines, which use a centralized parameter server to synchronize parameters between them. Experiments on three datasets demonstrate the efficiency and effectiveness of our framework, at previously-unseen problem scales with better accuracy than previous methods. We believe our work is the first to address distance metric learning at such large scales, in an intelligent distributed fashion.

%

\bibliography{dml}
\bibliographystyle{iclr2015}

\end{document}